\documentclass[10pt,journal,compsoc]{IEEEtran}
\ifCLASSOPTIONcompsoc
  \usepackage[nocompress]{cite}
\else
  \usepackage{cite}
\fi
\ifCLASSINFOpdf
\else
\fi

\usepackage{makeidx}  
\usepackage{url}
\usepackage{amsmath}
\usepackage{amssymb}
\usepackage{latexsym}
\usepackage{subfigure}
\usepackage{graphicx}
\usepackage{multirow}
\usepackage{color}
\usepackage{hyperref}
\usepackage{marginnote}
\usepackage{lipsum}

\definecolor{red}{rgb}{0.1,0.1,0.8}
\definecolor{blue}{rgb}{0,0,0.8}
\definecolor{green}{rgb}{0,0.4,0}

\newcommand{\change}[2]{}
\newcommand{\lchange}[2]{}
\newcommand{\changedtext}{}
\newcommand{\changed}[3]{#3}
\newcommand{\lchanged}[3]{#3}

\usepackage{algorithm}
\usepackage{algorithmic}
\usepackage{caption}
\usepackage[utf8]{inputenc} 
\usepackage[T1]{fontenc}    
\usepackage{hyperref}       
\usepackage{url}            
\usepackage{booktabs}       
\usepackage{amsfonts}       
\usepackage{nicefrac}       
\usepackage{microtype}      
\usepackage{graphicx}
\usepackage{subfigure}
\usepackage{amsmath}
\usepackage{color}
\usepackage{bm}
\usepackage{multirow}
\usepackage{float}

\begin{document}
\clearpage

\twocolumn
\pagenumbering{arabic}
\setcounter{page}{1}
\setcounter{table}{0}

\title{Universal Adversarial Attack on Attention and the Resulting Dataset DAmageNet}

\author{Sizhe~Chen, Zhengbao~He, Chengjin~Sun, Jie Yang, and Xiaolin~Huang,~\IEEEmembership{Senior~Member,~IEEE}
\IEEEcompsocitemizethanks{\IEEEcompsocthanksitem S.~Chen, Z.~He, C.~Sun, J. Yang, and  X.~Huang are with Department of Automation, and the Institute of Medical Robotics, Shanghai Jiao Tong University, and also with the MOE Key Laboratory of System Control and Information Processing, 800 Dongchuan Road, Shanghai, 200240, P.R. China. (e-mails:\{sizhe.chen, lstefanie, sunchengjin, jieyang, xiaolinhuang\}@sjtu.edu.cn)\protect\\
\IEEEcompsocthanksitem Corresponding author: Xiaolin Huang.}
\thanks{Manuscript received 2020.}}

\markboth{IEEE Transactions on Pattern Analysis and Machine Intelligence} 
{Chen \MakeLowercase{\textit{et al.}}: Universal Adversarial Attack on Attention and the Resulting Dataset DAmageNet}

\IEEEtitleabstractindextext{
\begin{abstract}
Adversarial attacks on deep neural networks (DNNs) have been found for several years. However, the existing adversarial attacks have high success rates only when the information of the victim DNN is well-known or could be estimated by the structure similarity or massive queries. In this paper, we propose to \emph{Attack on Attention} (AoA), a semantic property commonly shared by DNNs. AoA enjoys a significant increase in transferability when the traditional cross entropy loss is replaced with the attention loss. Since AoA alters the loss function only, it could be easily combined with other transferability-enhancement techniques and then achieve SOTA performance. We apply AoA to generate 50000 adversarial samples from ImageNet validation set to defeat many neural networks, and thus name the dataset as \emph{DAmageNet}. 13 well-trained DNNs are tested on DAmageNet, and all of them have an error rate over 85\%. Even with defenses or adversarial training, most models still maintain an error rate over 70\% on DAmageNet. DAmageNet is the first universal adversarial dataset. It could be downloaded freely and serve as a benchmark for robustness testing and adversarial training.
\end{abstract}
\begin{IEEEkeywords}
adversarial attack, attention, transferability, black-box attack, DAmageNet.
\end{IEEEkeywords}}

\maketitle
\IEEEdisplaynontitleabstractindextext
\IEEEpeerreviewmaketitle

\IEEEraisesectionheading{\section{Introduction}\label{introduction}}
\IEEEPARstart{D}{eep} neural networks (DNNs) have grown into the mainstream tools in many fields, thus, their vulnerability has attracted much attention in the recent years. An obvious example is the existence of adversarial samples \cite{akhtar2018threat}, which are quite similar with the clean ones, but are able to cheat the DNNs to produce incorrect predictions in high confidence. Various attack methods to craft adversarial samples have been proposed, such as FGSM \cite{goodfellow2014explaining}, C\&W \cite{carlini2017towards}, PGD \cite{madry2017towards}, Type I \cite{tang2019adversarial} and so on. Generally speaking, when the victim network is exposed to the attacker, one can easily achieve efficient attack with a very high success rate.

Although white-box attacks can easily cheat DNNs, the current users actually do not worry about them, since it is almost impossible to get the complete information including the structure and the parameters of the victim DNNs. If the information is kept well, one has to use black-box attack, which can be roughly categorized into query-based approaches \cite{cheng2019improving, ilyas2018prior, guo2019subspace} and transfer-based approaches \cite{papernot2017practical, moosavi2017universal, dong2019evading}. The former one is to estimate the gradient by querying the victim DNNs. However, until now, the existing query-based attacks still need massive queries, which can be easily detected by the defense systems. Transfer-based attacks rely on the similarity between the victim DNN and the attacked DNN, which serves as the \emph{surrogate model} in a black-box attack, in the attacker's hands. It is expected that white-box attacks on the surrogate model can also invade the victim DNN. Although there are some promising studies recently \cite{dong2018boosting, xie2019improving, lin2019nesterov}, the transfer performance is not satisfactory and a high attack rate could be reached only when two DNNs have similar structures \cite{su2018robustness}, which however conflicts the aim of black-box attacks.

Black-box adversarial samples that are applicable to vast DNNs need to attack their common vulnerability. Since DNNs are imitating human's intelligence, although DNNs have different structures and weights, they may share similar semantic features. In this paper, we are focusing on the attention heat maps, on which different DNNs have similar results. By attacking the heat maps of one white-box DNN, we could make its attention lose focus and therefore fail in judgement. In fact, some works have been aware of the importance of attention and put the change of heat map as an evidence of successful attacks, see, e.g.  \cite{dong2019evading, zhang2019interpreting}. But none of them includes the attention in loss. In our study, we develop an \emph{Attack on Attention} (AoA). AoA has a good white-box attack performance. More importantly, there is high similarity in attention across different DNNs, making AoA highly transferable: replacing the cross-entropy loss by AoA loss increases the transferability by 10\% to 15\%. Combined with some existing transferability-enhancement methods, AoA achieves a state-of-the-art performance, e.g. over 85\% transfer rate on all 12 black-box popular DNNs in numerical experiments.

\begin{figure}[htbp]
	\centering
	\includegraphics[width=\hsize]{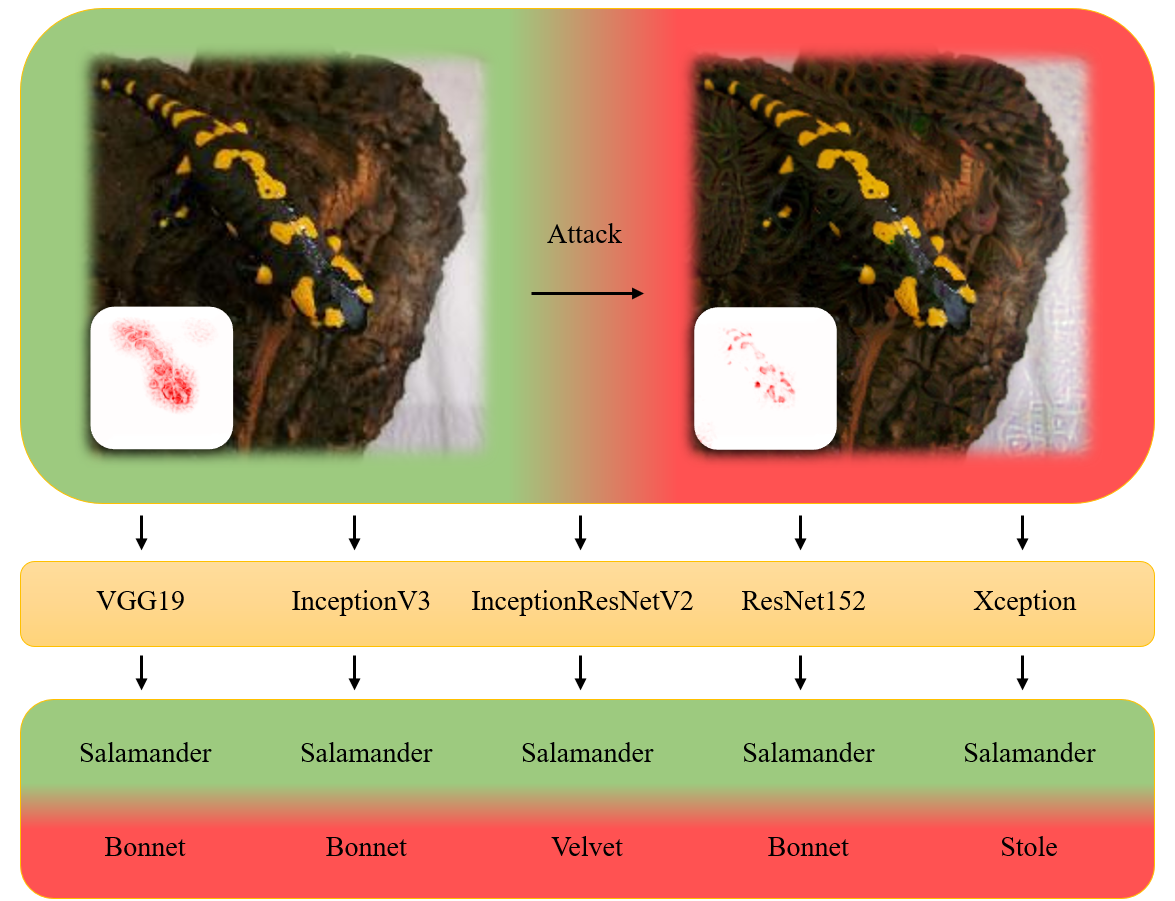}
	\caption{AoA adversarial sample and its attention heat map (calculated by DenseNet121). The original sample (in ImageNet: image n01629819\_15314.JPEG, class No.25) is shown on the left. All well-trained DNNs (listed in the first row) correctly recognize this image as a salamander. The right image is the generated adversarial sample by AoA. The difference between the two images is slight, however, the heat map shown in lower left corner changes a lot, which fools all the listed DNNs to incorrect predictions, as shown in the bottom row.}
	\label{intro}
\end{figure}

Here, we first illustrate one example in Fig. \ref{intro}. The original image is a "salamander" in ImageNet \cite{deng2009imagenet}. By attacking the attention, we generate an adversarial sample, which looks very similar to the original one but with a scattered heat map (in the lower left corner), leading to misclassification. The attack is carried out on VGG19 \cite{simonyan2014very} but other well-trained DNNs on ImageNet also make wrong predictions.

Since AoA is for common vulnerabilities of DNNs, we successfully generate 50000 adversarial samples that can cheat many DNNs, of which the error rates increase to over $85\%$. We provide these samples in the dataset named as \emph{DAmageNet}. DAmageNet is the first dataset that provides black-box adversarial samples. Those images \emph{DAmage} many neural networks without any knowledge or query. But the aim is not to really damage them, but to point out the weak parts of neural networks and thus those samples are valuable to improve the neural networks by adversarial training \cite{ganin2016domain, shrivastava2017learning}, robustness certification \cite{sinha2017certifiable}, and so on.

The rest of this paper is organized as follows. In Section \ref{related}, we will briefly introduce adversarial attack, especially black-box attack, attention heat map, and several variants of ImageNet. The Attack on Attention is described in detail in Section \ref{method}. Section \ref{experiment} evaluates the proposed AoA along with other attacks and defenses and presents the DAmageNet. In Section \ref{conclusion}, a conclusion is given to end this paper.

\section{Related Work}\label{related}
\subsection{Adversarial attack and its defense}\label{related-defense}
Adversarial attacks \cite{szegedy2013intriguing} reveal the weakness of DNNs by cheating it with adversarial samples, which differ from original ones with only a slight perturbation. In the humans' eyes, the adversarial samples do not differ from the original ones, but well-trained networks make false predictions on them in high confidence. The adversarial attack can be expressed as below,
\begin{eqnarray*}
\begin{split}
{\text { find }} & {\Delta x} \\ {\text { s.t. }} & {f(x) \neq f(x+\Delta x)} \\ {} & {\|\Delta x\| \leq \varepsilon},
\end{split}
\end{eqnarray*}
\lchanged{M01}{\link{R0.0} \link{R3.2}}{where a neural network $f$ predicts differently on the clean sample and the adversarial sample. Even their difference is imperceivable, i.e., $\Delta x$ is restricted by $||\cdot||$, which could be the $\ell_1$-, $\ell_2$- or $\ell_\infty$-norm.}

When training a DNN, one updates the weights of the network by the gradients to minimize a training loss. While in adversarial attacks, one alters the image to increase the training loss. Based on this basic idea, there have been many variants on attacking spaces and crafting methods.

For the space to be attacked, most of the existing methods directly conduct attack in the image space \cite{goodfellow2014explaining, moosavi2016deepfool, su2019one}. It is also reasonable to attack the feature vector in the latent space \cite{song2018constructing, tang2019adversarial} or the encoder/decoder \cite{baluja2017adversarial, han2019once}. Attack on feature space may produce unique perturbation unlike random noise.

Adversarial attacks could be roughly categorized as gradient-based \cite{goodfellow2014explaining, madry2017towards} and optimization-based methods \cite{szegedy2013intriguing, carlini2017towards}. Gradient-based methods search in the gradient direction and the magnitude of perturbation is restricted to avoid a big distortion. Optimization-based methods usually consider the magnitude restriction in the objective function. For both, the magnitude could be measured by the $\ell_1$, $\ell_2$, $\ell_\infty$-norm or other metrics.

To secure the DNN, many defense methods have been proposed to inhibit the adversarial attack. Defense can be achieved by adding adversarial samples to the training set, which is called adversarial training \cite{miyato2016adversarial, sankaranarayanan2018regularizing, zhang2019you}. It is very effective, but consumes several-fold time. Another technique is to design certain blocks in the network structure to prevent attacks or detect adversarial samples \cite{liao2018defense, xie2019feature}. Attack can also be mitigated by preprocessing images before input to the DNN \cite{liu2018feature, prakash2018deflecting, mustafa2019image}, which does not require modification on the pre-trained network.

\subsection{Black-box attack}
When the victim DNNs are totally known, the attacks mentioned above have high success rates. However, it is almost impossible to have access to the victim model in real-world scenarios and thus black-box attacks are required \cite{papernot2016transferability, brendel2017decision, ilyas2018black}. Black-box attacks rely on either query \cite{cheng2019improving, ilyas2018prior} or transferability \cite{papernot2016transferability, papernot2017practical}.

For the query-based approach, the attacker adds a slight perturbation to the input image and observes the reaction of the victim model. By a series of queries, the gradients could be roughly estimated and then one can conduct the attack in the way similar to white-box cases. To decide on the attack direction, attackers adopt methods including Bayes optimization \cite{ru2020bayesopt}, evolutional algorithms \cite{laurent2019yet}, meta learning \cite{du2019query} etc. Since the practical DNNs are generally very complicated, good estimation of the gradients needs a massive number of queries, leading to an easy detection by the model owner.

For the transfer-based approach, one conducts white-box attack in a well-designed surrogate model and expects that the adversarial samples remain aggressive to other models. The underlying assumption is that the distance between decision boundaries across different classes is significantly shorter than that across different models \cite{papernot2016transferability}. Although a good transfer rate has been recently reported in \cite{xie2019improving, dong2018boosting, lin2019nesterov, wu2019skip}, it is mainly for models in the same family, e.g., InceptionV3 and InceptionV4, or models with the same blocks, e.g., residual blocks \cite{su2018robustness}. Until now, cross-family transferability of adversarial samples with small perturbations is limited and there is no publicly available dataset of that.

\subsection{Attention heat map}
In making judgements, humans tend to concentrate on certain parts of an object and allocate attention efficiently. This attention mechanism in human intelligence has been exploited by researchers. In recent studies, methods in natural language process have benefited from the attention mechanism a lot \cite{vaswani2017attention}. In computer vision, the same idea has been applied and becomes an important component in DNNs, especially in industrial applications \cite{samek2019explainable}.

To attack on attention, we need to calculate the pixel-wise attention heat map, for which network visualization methods \cite{zhou2016learning, lin2013network} are applicable. Forward visualization adopts the intuitive idea to obtain the attention by observing the changes in the output caused by changes in the input. The input can be modified by noise \cite{zhou2014object}, masking \cite{zeiler2014visualizing}, or perturbation \cite{zhou2015predicting}. However, these methods consume much time and may introduce randomness.

In contrast, backward visualization \cite{simonyan2013deep, zeiler2014visualizing, springenberg2014striving} obtains the heat map by calculating the relevance between adjacent layers from the output to the input. The layer-wise attention is obtained by the attention in the next layer and the network weights in this layer. Significant works include Layer-wise Relevance Propagation (LRP) \cite{bach2015pixel}, Contrastive LRP (CLRP) \cite{gu2018understanding} and Softmax Gradient LRP (SGLRP) \cite{iwana2019explaining}. These methods extract the high-level semantic attention features for the images from the perspective of the network and make DNNs more interpretable and explainable.

\subsection{ImageNet and its variants}
To demonstrate and evaluate our attack, we will modify images from ImageNet as other transfer attacks \cite{xie2019improving, dong2018boosting, lin2019nesterov, wu2019skip}. ImageNet is a large-scale dataset \cite{deng2009imagenet}, which contains images of 1000 classes and each has 1300 well-chosen samples. ImageNet Large Scale Visual Recognition Challenge (ILSVRC) has encouraged a lot of mile-stone works \cite{simonyan2014very, he2016deep, huang2017densely}. Recently, many interesting variants of ImageNet have been developed, including ImageNet-A \cite{hendrycks2019natural}, ObjectNet \cite{barbu2019objectnet}, ImageNet-C, and ImageNet-P \cite{hendrycks2019benchmarking}.

ImageNet-A contains real-world images in ImageNet classes, and they are able to mislead many classifiers to output false predictions. ObjectNet also includes natural images that well-trained models in ImageNet cannot distinguish. Objects in ObjectNet have random backgrounds, rotations and viewpoints. ImageNet-C is produced by adding 15 diverse corruptions. Each type of corruptions has 5 levels from the lightest to the severest. ImageNet-P is designed from ImageNet-C and differs from it in possessing additional perturbation sequences, which are not generated by attack but by image transformations.

The datasets mentioned above are valuable for testing and improving the network generalization capability, but DAmageNet is for the robustness. In other words, samples in the above datasets differ from the samples in ImageNet and the low accuracy is due to the poor generalization. In DAmageNet, the samples are quite similar to the original ones in ImageNet and the low accuracy is due to the over-sensitivity of DNNs.

\section{Attack on Attention (AoA)}\label{method}
\begin{figure}[htbp]
	\centering
	\includegraphics[width=\hsize]{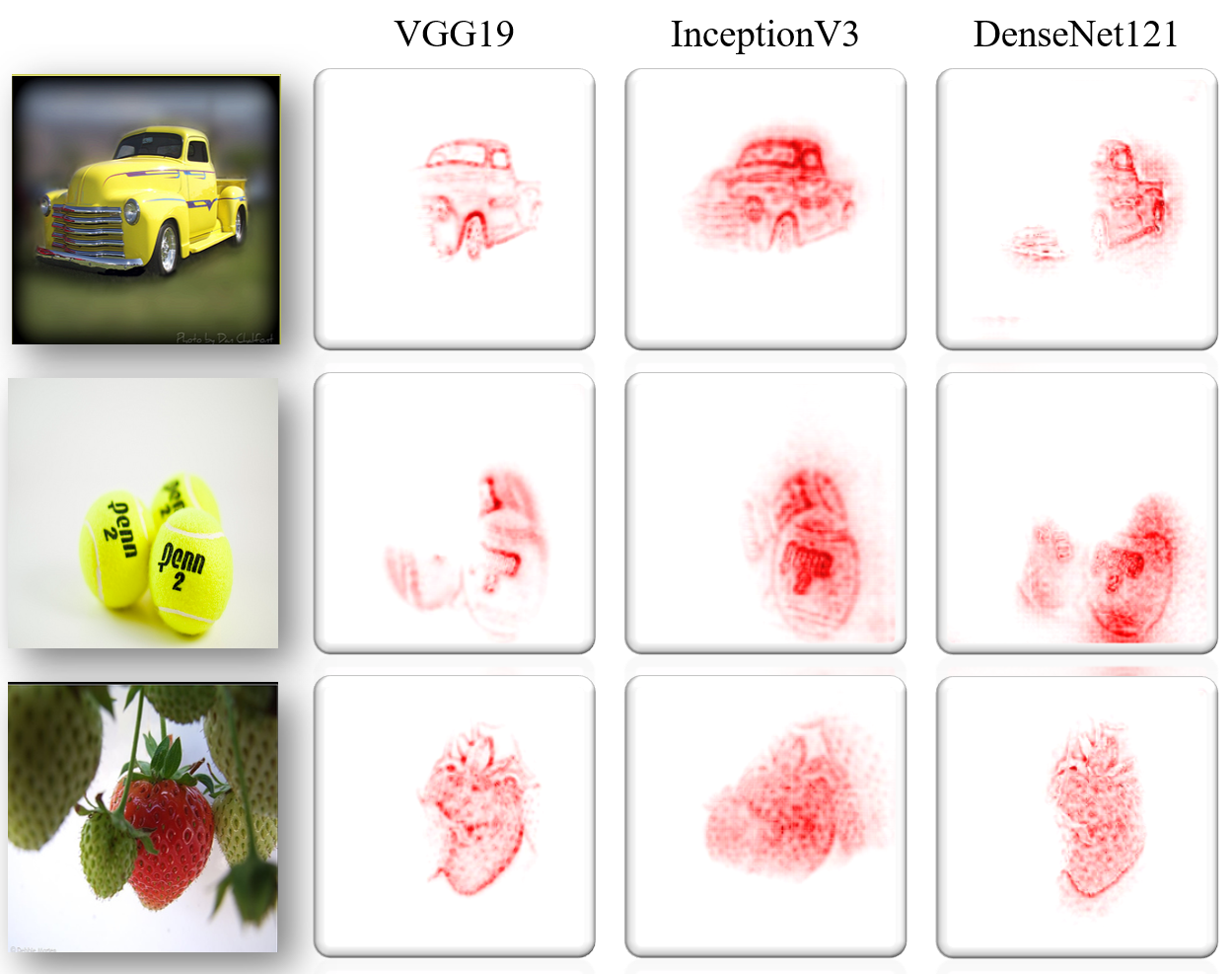}
	\caption{Attention heat maps for VGG19 \cite{simonyan2014very}, InceptionV3 \cite{szegedy2016rethinking}, DenseNet121 \cite{huang2017densely}, which are similar even the architectures are different.}
	\label{heatmaps}
\end{figure}

To pursue high transferability for black-box attacks, we need to find common vulnerabilities and attack semantic features shared by different DNNs. Attention heat maps for three images are illustrated in Fig. \ref{heatmaps}, where the pixel-wise heat maps show how the input contributes to the prediction. Even with different architectures, the models have similar attention. Inspired by the similarity across different DNNs, we propose to Attack on Attention (AoA). Different to the existing methods that focus on attacking the output, AoA aims to change the attention heat map.

\begin{figure}[htbp]
	\centering
	\includegraphics[width=\hsize]{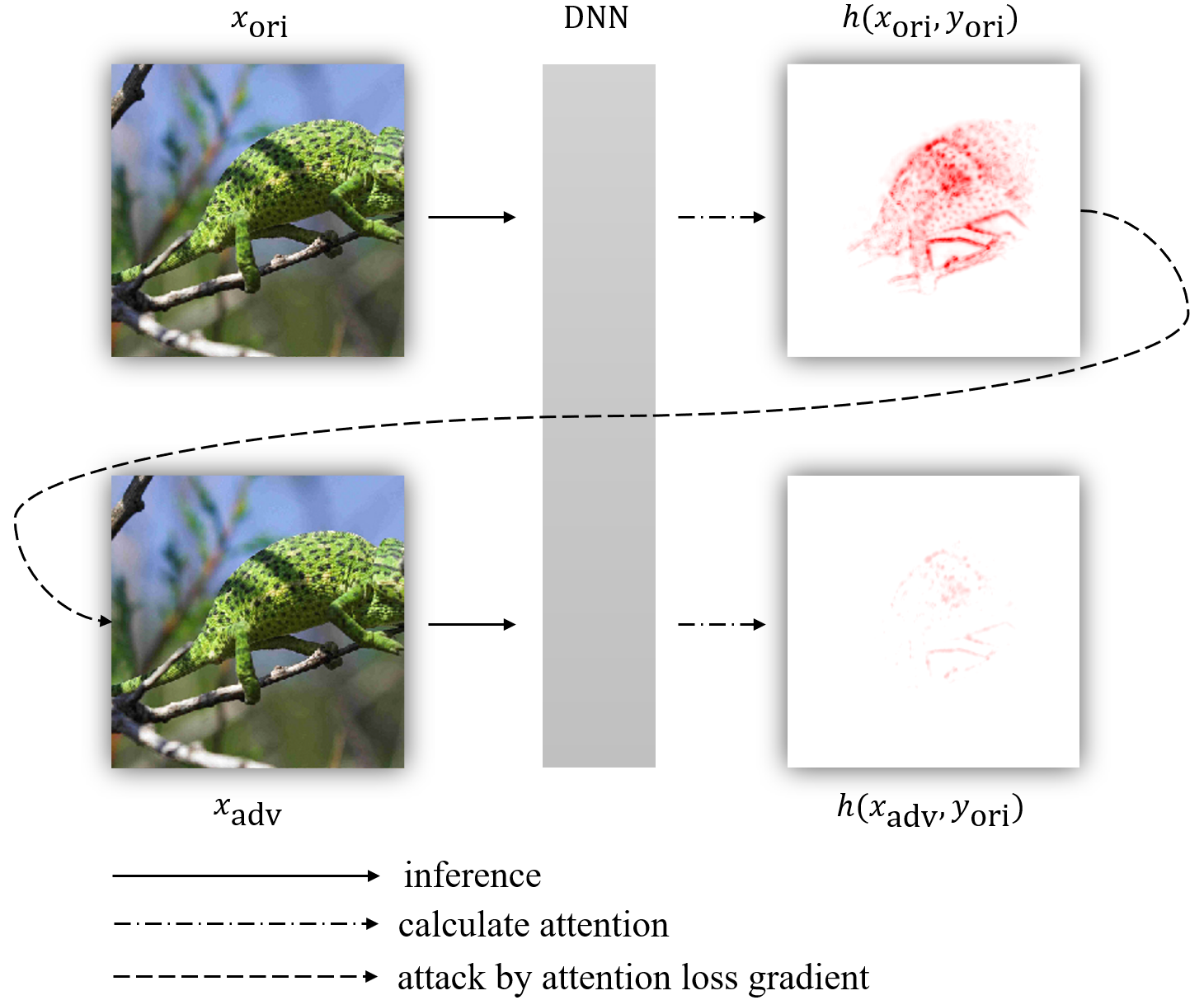}
	\caption{The design of AoA. AoA calculates the attention heat map by SGLRP after inference. The gradient from the heat map back-propagates to the input and updates the sample iteratively. By suppressing the attention heat map value, one can change the network decision by fooling its focus. Constantly doing this, the produced adversarial sample could beat several black-box models.}
	\label{loss}
\end{figure}

Let $h(x, y)$ stand for the attention heat map for the input $x$ and a specified class $y$. $h(x, y_\mathrm{ori})$ is a tensor with the dimension consistent to $x$. The basic idea of AoA is to shift the attention away from the original class, e.g. decrease the heat map for the correct class $y_\mathrm{ori}$, as illustrated in Fig. \ref{loss}. In this paper, we utilize SGLRP \cite{iwana2019explaining} to calculate the attention heat map $h(x, y)$, which is good at distinguishing the attention for the target class from the others. There exist of course many other techniques for obtaining the heat map to attack, as long as $h(x, y)$ and its gradient on $x$ could be effectively calculated.

\begin{table*}[htbp]
	\caption{Transfer Rate from ResNet50 to Other Neural Networks}
	\centering
	\begin{tabular}{c|ccccccc}
\toprule
Loss/Method & DN121 \cite{huang2017densely} & VGG19 \cite{simonyan2014very} & RN152 \cite{he2016deep} & IncV3 \cite{szegedy2016rethinking} & IncRNV2 \cite{szegedy2017inception} & Xception \cite{chollet2017xception} & NASNetL \cite{zoph2018learning} \\ \midrule
CW \cite{carlini2017towards} & 66.6$\pm$1.24\% & 54.2$\pm$4.27\% & 47.3$\pm$4.69\% & 39.6$\pm$2.92\% & 37.9$\pm$4.77\% & 37.4$\pm$2.67\% & 28.8$\pm$2.58\% \\
PGD \cite{madry2017towards} & 67.8$\pm$1.83\% & 54.2$\pm$2.56\% & 46.8$\pm$3.71\% & 38.7$\pm$2.25\% & 35.6$\pm$4.21\% & 37.4$\pm$4.08\% & 28.4$\pm$3.17\% \\ \midrule
$L_\mathrm{supp}(x)$ & 66.8$\pm$3.37\% & 57.2$\pm$3.96\% & 54.8$\pm$2.50\% & 43.9$\pm$2.78\% & 41.6$\pm$1.66\% & 40.9$\pm$2.60\% & 33.0$\pm$2.53\% \\
$L_\mathrm{dstc}(x)$ & 67.1$\pm$4.04\% & 56.5$\pm$2.28\% & 55.5$\pm$4.15\% & 45.4$\pm$3.77\% & 40.0$\pm$1.82\% & 41.6$\pm$4.07\% & 31.0$\pm$2.17\% \\
$L_\mathrm{bdry}(x)$ & 50.2$\pm$5.26\% & 49.8$\pm$4.39\% & 44.0$\pm$4.05\% & 34.1$\pm$3.34\% & 32.9$\pm$3.22\% & 31.7$\pm$1.86\% & 21.7$\pm$1.29\% \\
$L_\mathrm{log}(x)$ & 74.9$\pm$3.48\% & 64.2$\pm$4.13\% & 59.2$\pm$4.71\% & 50.1$\pm$2.69\% & 46.2$\pm$3.39\% & 48.0$\pm$4.87\% & 36.3$\pm$3.74\% \\
\midrule
$L_\mathrm{AoA}(x)$ & \textbf{78.7$\pm$2.54\%} & \textbf{64.9$\pm$2.01\%} & \textbf{63.9$\pm$1.98\%} & \textbf{53.3$\pm$2.27\%} & \textbf{48.9$\pm$2.65\%} & \textbf{50.9$\pm$3.01\%} & \textbf{41.0$\pm$2.00\%} \\
\bottomrule
    \end{tabular}
	\label{toy}
\end{table*}

There are several potential ways to change the attention heat maps.
\begin{enumerate}\label{scheme}
    \item Suppress the magnitude of attention heat maps for the correct class $h(x, y_\mathrm{ori})$: When the network attention degree on the correct class decreases, attention for other classes would increase and finally exceed the correct one, which leads the model to seek for information on other classes rather than the correct one and thus make an incorrect prediction. We call this design as the following \emph{suppress loss,
        \begin{eqnarray*}
        L_\mathrm{supp}(x) = \|h(x, y_\mathrm{ori})\|_1,
        \end{eqnarray*}}
        where $\|\cdot\|_1$ stands for the componentwise $\ell_1$-norm.
    \item Distract the focus of $h(x, y_\mathrm{ori})$: It could be expected that when the attention is distracted from the original regions of interest, the model may lose its capability for prediction. In this case, we do not require the network to focus on information of any incorrect class, but lead it to concentrate on irrelevant regions of the image. The loss could be expressed as the following \emph{distract loss},
        \begin{eqnarray*}
        L_\mathrm{dstc}(x) = -\left\|\frac{h(x, y_\mathrm{ori})}{max(h(x, y_\mathrm{ori}))} - \frac{h(x_\mathrm{ori}, y_\mathrm{ori})}{max(h(x_\mathrm{ori}, y_\mathrm{ori}))}\right\|_1.\\
        \end{eqnarray*}
        Here, self-normalization is conducted to eliminate the influence of attention magnitude.
    \item Decrease the gap between $h(x, y_\mathrm{ori})$ and $h(x, y_\mathrm{sec}(x))$, the heat map for the second largest probability: If the attention magnitude for the second class exceeds that for the correct class, the network would focus more on information about the false prediction, which is inspired by CW attack \cite{carlini2017towards}. We call it \emph{boundary loss} and take the following formulation,
        \begin{eqnarray*}
        L_\mathrm{bdry}(x) = \|h(x, y_\mathrm{ori})\|_1-\|h(x, y_\mathrm{sec}(x))\|_1.
        \end{eqnarray*}
        The values of attention heat maps vary a lot for different models, so the self-normalization may improve the transferability of adversarial samples. Therefore, rather than $L_\mathrm{bdry}$, we can also consider the ratio between $h(x, y_\mathrm{ori})$ and $h(x, y_\mathrm{sec}(x))$, resulting the following \emph{logarithmic boundary loss}
        \begin{eqnarray*}
        \begin{split}
        L_\mathrm{log}(x) = \log(\|h(x, y_\mathrm{ori})\|_1) - \log(\|h(x, y_\mathrm{sec}(x))\|_1).
        \end{split}
        \end{eqnarray*}
\end{enumerate}

\begin{figure}[htbp]
	\centering
	\includegraphics[width=0.95\hsize]{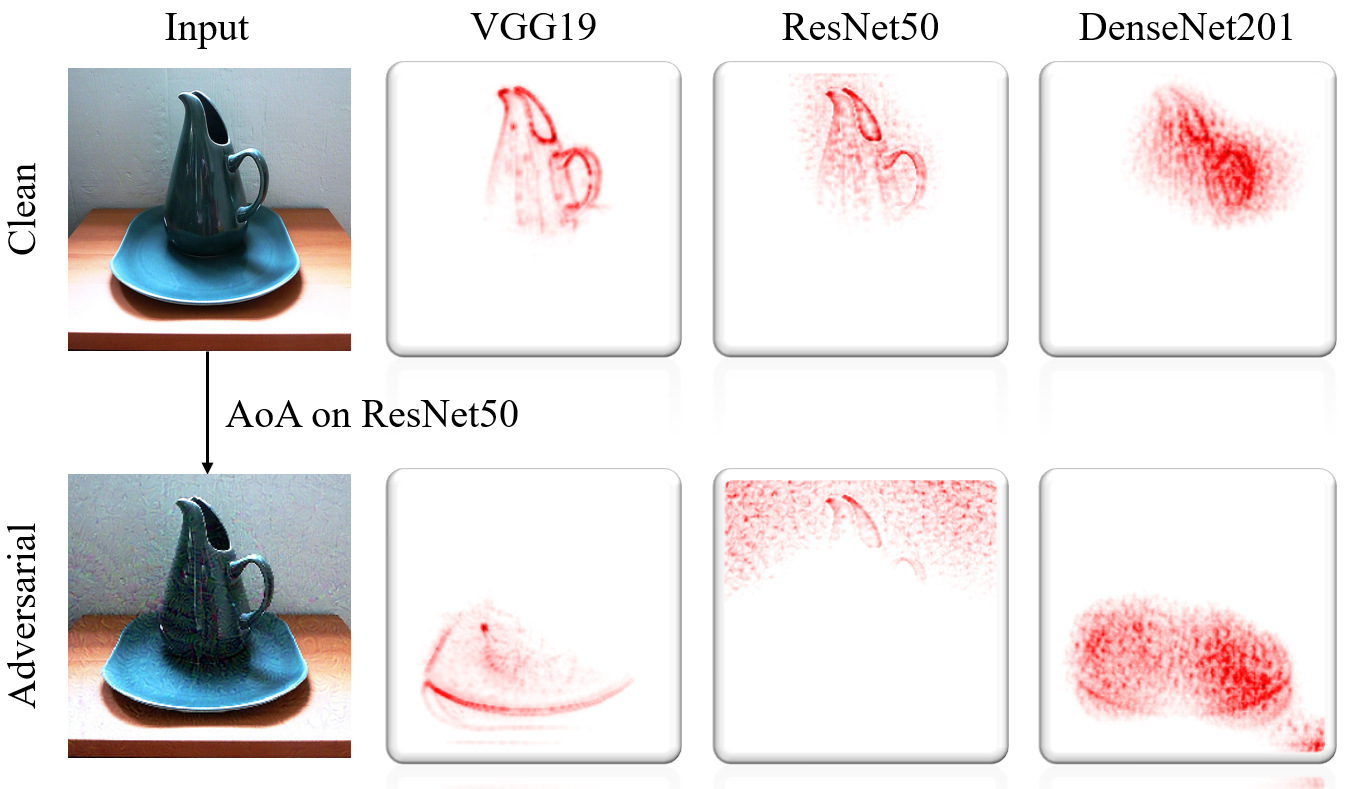}
	\caption{Minimizing $L_\mathrm{dstc}$ distracts the attention from the correct ROI to irrelevant regions and similar distraction could be observed for different networks.}
	\label{dstc}
\end{figure}

Now let us illustrate the attack result on the attention heat map by distract loss. In Fig. \ref{dstc}, a clean sample is drawn together with its heat maps away from its original class.
Aiming at ResNet50 \cite{he2016deep}, we minimize $L_\mathrm{dstc}$ and successfully change the heat map such that the attention is distracted to irrelevant regions (the second right column at the bottom). This common property shared by the attention in different DNNs makes the attack transferable, which is the motivation of attack on attention. The generated adversarial sample
is shown in the leftmost in the bottom, which is incorrectly
recognized by all the DNNs in Fig. \ref{dstc}. Additionally, we could see that the heat map for VGG19 is much clearer, which might explain the high transferability of its adversarial samples as shown later and in \cite{su2018robustness}.

The transferability across different DNNs could be observed not only for the $L_\mathrm{dstc}$ but also for the other attention-related losses. To compare the above losses' attack performance, we attack on ResNet50 \cite{he2016deep} and feed the adversarial samples to other DNNs (see the setting in Section \ref{experiment} for details). Two attacks on classification loss, namely CW and PGD, are also compared as the baseline.
The white-box attack success rates. i.e., the error rates of ResNet50, are all near $100\%$ but attacks by different losses have different transferability performance, which is reported in Table \ref{toy}. The suppress loss and the distract loss have a better transferability than PGD and CW. The logarithmic boundary loss is the best and is hence chosen as the attack target. Moreover, attack on attention could be readily combined with the existing attack on prediction (the cross entropy loss attacked in PGD, denoted by $L_\mathrm{ce}$), resulting in the following \emph{AoA loss},
\begin{eqnarray}\label{aoaloss}
L_\mathrm{AoA}(x) = L_\mathrm{log}(x) - \lambda  L_\mathrm{ce}(x, y_\mathrm{ori}),
\end{eqnarray}
where $\lambda$ is a trade-off between the attack on attention and cross entropy. In this paper, $\lambda=1000$ is suggested such that the two items have similar variance for different inputs. The combination further increases the transferability, as shown in Table \ref{toy}.

Basically, the adversarial samples are generated in an update process by minimizing the AoA loss $L_\mathrm{AoA}$. Specifically, set $x^0_\mathrm{adv} = x_\mathrm{ori}$ and the update procedure could be generally described as the following
\begin{eqnarray}\label{update}
\begin{split}
x^{k+1}_\mathrm{adv} & = \text{clip}_\varepsilon\left(x^{k}_\mathrm{adv} - \alpha \frac{g(x^{k}_\mathrm{adv})}{||g(x^{k}_\mathrm{adv})||_1/N} \right), \\
g(x) &= \frac{\partial L_\mathrm{AoA}(x)}{\partial x}.
\end{split}
\end{eqnarray}
The gradient $g$ is normalized by its average $\ell_1$-norm, i.e., $||g(x_k)||_1/N$, where $N$ is the size of the image. Further, to keep the perturbations invisible, we restrict our attack by the distance from the original clean sample 
such that the $\ell_\infty$ distance does not exceed $\varepsilon$. AoA is different from other attacks merely on the loss. Therefore, transferability-enhancement techniques developed for directly attacking prediction are also applicable to AoA. In fact, with optimization modification \cite{dong2018boosting} or input modification \cite{xie2019improving, dong2019evading, lin2019nesterov}, the transfer performance of AoA gets further improved, as numerically verified in Section \ref{experiment-trans}. \lchanged{M02}{\link{R0.0} \link{R3.3}}{The procedure of AoA is summarized in Algorithm \ref{alg}.}

\renewcommand{\algorithmicrequire}{\textbf{Input:}}
\renewcommand{\algorithmicensure}{\textbf{Output:}}

\begin{algorithm}[htbp]
  \caption{\changedtext{Attack on Attention}}
  \label{alg}
  \begin{algorithmic}[1]
    \REQUIRE{AoA loss $L_\mathrm{AoA}(x)$, origin sample $x_{\mathrm{ori}}$, $\ell_{\infty}$-norm bound $\epsilon$, RMSE threshold $\eta$, attack step length $\alpha$.}

    \ENSURE{adversarial sample $x_{\mathrm{adv}}$}

    \STATE $x_{\mathrm{adv}}^{0} \gets x_{\mathrm{ori}}$

    \STATE $N \gets height \times width \times channel$  of   $x_{\mathrm{ori}}$
    \STATE $k \gets 0$

    \WHILE {$RMSE(x_{\mathrm{ori}}, x_{\mathrm{adv}}^{k}) < \eta $}
        \STATE $g = \frac{\partial L_{AoA}(x_{\mathrm{adv}}^{k})}{\partial x_{\mathrm{adv}}^{k}} \quad \quad \quad \quad \quad \quad \quad \quad \quad \quad \quad ~ ~ ~  \quad \quad  \quad \star $

        \STATE $x_{\mathrm{adv}}^{k+1}= \mathrm{clip}_{\epsilon}(x_{\mathrm{adv}}^{k} - \alpha \cdot \frac{g}{||g||_{1}/N}) \quad  \quad ~ \quad\quad \quad \quad \quad \star \star $

        \STATE $k = k+1$
    \ENDWHILE
    \RETURN ${x_{\mathrm{adv}}^{k}}$

    \STATE
    \STATE $\star ~~$: could be modified for DI \cite{xie2019improving},SI \cite{lin2019nesterov} enhancement.
    \STATE $\star \star$: could be modified for MI \cite{dong2018boosting},TI \cite{dong2019evading} enhancement.
    \end{algorithmic}
\end{algorithm}

Because of its good transferability on attention heat maps, AoA could be used for the black-box attack. The basic scheme is to choose a white-box DNN, which serves as the surrogate model for black-box attacks, to attack 
by updating (\ref{update}). The generated adversarial samples tend to be aggressive to other black-box victim models.

\section{Experiments}\label{experiment}
In this section, we will evaluate the performance of our Attack on Attention, especially its black-box attack capability compared to other SOTA methods. Since AoA is a very good black-box attack, it provides adversarial samples that can defeat many DNNs in a zero-query manner. These samples are collected in the dataset DAmageNet. This section will also introduce DAmageNet and report the performance of different DNNs on it. We further test the AoA performance under several defenses and find that AoA is the most aggressive method in almost all the cases.

\subsection{Setup}
The experiments for AoA are conducted on ImageNet \cite{deng2009imagenet} validation set. For attack and test, several well-trained models in Keras Applications \cite{chollet2015keras} are used, including VGG19 \cite{simonyan2014very}, ResNet50 \cite{he2016deep}, DenseNet121 \cite{huang2017densely}, InceptionV3 \cite{szegedy2016rethinking} and so on. We also use other adversarial-trained models (not by AoA, indicated by underline).  For preprocessing, Keras preprocessing function, central cropping, and resizing (to 224) are used. The experiments are implemented in TensorFlow \cite{tensorflow2015-whitepaper}, Keras \cite{chollet2015keras} with 4 NVIDIA GeForce RTX 2080Ti GPUs.

For the attack performance, we care about two aspects: the success/transfer rate of attack and how large the image is changed. Denote the generated adversarial sample as $x_\mathrm{adv}$. The change from its corresponding original image $x_\mathrm{ori}$ could be measured by the Root Mean Squared Error (RMSE) in each pixel: $d\left(x_\mathrm{adv}, x_\mathrm{ori}\right)=\sqrt{\|x_{\mathrm{adv}}-x_{\mathrm{ori}}\|_2^{2} / N}$. In the experiments, 200 images are randomly selected
from ImageNet validation set and the samples incorrectly predicted by the victim model are skipped as the same setting in \cite{su2018robustness}. Experiments are repeated 5 times and the overall performance on 1000 samples is reported. All the compared attacks will be fairly stopped when RMSE exceeds $\eta=7$ and the perturbation is bounded by $\varepsilon=0.1*255$. In this way, the number of iterations is about 10 with step size $\alpha=2$ as the setting of \cite{wu2019skip} and other literatures. We alter $\alpha=0.5$ for MI \cite{dong2018boosting} \changed{M03}{\link{R0.0} \link{R1.1}}{based on numerical experiments}.

\subsection{Transferability of AoA}\label{experiment-trans}
We first compare AoA with popular attacks CW \cite{carlini2017towards} and PGD \cite{madry2017towards}, which aim at classification losses. Specifically, CW uses the hinge loss and PGD uses the cross entropy loss. For CW, a gradient-based update is applied to keep the perturbation small. We carefully tune their parameters, resulting in a better transferability than reported in \cite{su2018robustness}.

We use AoA, CW, and PGD to attack different neural networks, and then feed the generated adversarial samples to different models. The average error rates are reported in Table \ref{aoaresult}. AoA, CW, and PGD all have a high white-box attack success rate but the transfer performance varies a lot, which depends on both the surrogate model and the victim model. But in all the tested situations, AoA achieves a better black-box attack performance.

The essential difference of AoA from CW/PGD is the attack target. The existing effort on improving attack transferability for CW/PGD is mainly on modifying the optimization process. For example, DI proposes to transform 4 times when calculating gradients with a probability \cite{xie2019improving}. \changed{M04}{\link{R0.0} \link{R1.2}}{TI translates the image for more transferable attack gradients \cite{dong2019evading}.}
MI tunes momentum parameter for boosting attacks \cite{dong2018boosting}. SI divides the sample by the power 2 for 4 times to calculate the gradient \cite{lin2019nesterov}. Those state-of-the-art transferability-enhancement methods could improve the performance for CW/PGD and are also applicable to AoA.

In Table \ref{aoaresult2}, we report the black-box attack performance when attacking ResNet50 with MI-DI, \changedtext{MI-TI}, and SI (all with the hyperparameters suggested by their inventors). We find that SI is very helpful and can prominently increase the error rate for PGD and CW. Applying SI in AoA, denoted as SI-AoA, achieves the highest transfer rate, which is significantly better than other state-of-the-art methods.

\begin{table*}[htbp]
	\caption{Error Rate (Top-1) of Different Attack Baselines}
	\centering
    \setlength{\tabcolsep}{3pt}{
	\begin{tabular}{l|c|cccccccc}
\toprule
Surrogate & Method & DN121 \cite{huang2017densely} & IncRNV2 \cite{szegedy2017inception} & IncV3 \cite{szegedy2016rethinking}  & NASNetL \cite{zoph2018learning} & RN152 \cite{he2016deep} & RN50 \cite{he2016deep} & VGG19 \cite{simonyan2014very} & Xception \cite{chollet2017xception}\\ \midrule
 & CW & 66.6$\pm$1.24\% & 37.9$\pm$4.77\% & 39.6$\pm$2.92\% & 28.8$\pm$2.58\% & 47.3$\pm$4.69\% & 100.0$\pm$0.00\% & 54.2$\pm$4.27\% & 37.4$\pm$2.67\% \\
RN50 \cite{he2016deep} & PGD & 67.8$\pm$1.83\% & 35.6$\pm$4.21\% & 38.7$\pm$2.25\% & 28.4$\pm$3.17\% & 46.8$\pm$3.71\% & \textbf{100.0$\pm$0.00\%} & 54.2$\pm$2.56\% & 37.4$\pm$4.08\% \\
 & AoA & \textbf{78.4$\pm$2.44\%} & \textbf{49.0$\pm$1.87\%} & \textbf{52.2$\pm$2.66\%} & \textbf{39.6$\pm$3.61\%} & \textbf{63.4$\pm$2.63\%} & 99.9$\pm$0.20\% & \textbf{65.6$\pm$2.82\%} & \textbf{51.1$\pm$2.18\%} \\ \midrule
 & CW & 100.0$\pm$0.00\% & 33.5$\pm$2.55\% & 39.5$\pm$1.67\% & 31.9$\pm$2.87\% & 39.6$\pm$2.85\% & 64.6$\pm$3.76\% & 53.2$\pm$3.93\% & 39.4$\pm$1.16\% \\
DN121 \cite{huang2017densely} & PGD & 100.0$\pm$0.00\% & 34.0$\pm$3.49\% & 41.7$\pm$2.38\% & 31.9$\pm$2.87\% & 41.5$\pm$3.21\% & 68.9$\pm$4.76\% & 55.5$\pm$2.28\% & 41.5$\pm$2.30\% \\
 & AoA & \textbf{100.0$\pm$0.00\%} & \textbf{46.1$\pm$2.91\%} & \textbf{53.5$\pm$3.46\%} & \textbf{46.1$\pm$2.44\%} & \textbf{55.0$\pm$2.77\%} & \textbf{76.7$\pm$2.29\%} & \textbf{64.6$\pm$2.18\%} & \textbf{52.1$\pm$2.15\%} \\ \midrule
 & CW & 31.0$\pm$1.95\% & 22.7$\pm$3.01\% & 100.0$\pm$0.00\% & 21.3$\pm$0.60\% & 26.1$\pm$3.62\% & 42.3$\pm$2.01\% & 40.7$\pm$3.34\% & 33.4$\pm$1.56\% \\
IncV3 \cite{szegedy2016rethinking} & PGD & 32.7$\pm$2.50\% & 24.2$\pm$2.89\% & 100.0$\pm$0.00\% & 21.3$\pm$1.91\% & 27.3$\pm$2.29\% & 45.3$\pm$1.17\% & 40.7$\pm$3.39\% & 33.7$\pm$3.22\% \\
 & AoA & \textbf{39.0$\pm$1.79\%} & \textbf{30.2$\pm$2.77\%} & \textbf{100.0$\pm$0.00\%} & \textbf{32.7$\pm$1.81\%} & \textbf{34.0$\pm$2.93\%} & \textbf{52.8$\pm$1.69\%} & \textbf{45.9$\pm$3.98\%} & \textbf{45.1$\pm$2.08\%} \\ \midrule
 & CW & 85.5$\pm$0.84\% & 62.0$\pm$1.67\% & 69.8$\pm$1.60\% & 62.7$\pm$1.21\% & 60.0$\pm$1.61\% & 77.8$\pm$2.04\% & 100.0$\pm$0.00\% & 68.0$\pm$2.39\% \\
VGG19 \cite{simonyan2014very} & PGD & 87.1$\pm$1.20\% & 64.1$\pm$2.03\% & 71.8$\pm$1.63\% & 63.9$\pm$1.77\% & 63.1$\pm$4.14\% & 82.5$\pm$2.63\% & 100.0$\pm$0.00\% & 71.9$\pm$0.97\% \\
 & AoA & \textbf{91.4$\pm$2.65\%} & \textbf{73.7$\pm$1.29\%} & \textbf{79.8$\pm$1.08\%} & \textbf{74.2$\pm$1.63\%} & \textbf{73.5$\pm$1.05\%} & \textbf{86.6$\pm$1.77\%} & \textbf{100.0$\pm$0.00\%} & \textbf{81.0$\pm$1.30\%} \\ \midrule
 & CW & 42.4$\pm$2.52\% & 36.2$\pm$2.32\% & 35.3$\pm$1.66\% & 25.6$\pm$2.24\% & 100.0$\pm$0.00\% & 57.7$\pm$0.81\% & 46.0$\pm$4.06\% & 31.9$\pm$1.77\% \\
RN152 \cite{he2016deep} & PGD & 42.7$\pm$3.19\% & 35.0$\pm$2.47\% & 34.9$\pm$2.96\% & 24.5$\pm$3.05\% & 98.1$\pm$0.97\% & 55.3$\pm$2.71\% & 43.6$\pm$3.61\% & 30.5$\pm$4.87\% \\
 & AoA & \textbf{55.9$\pm$2.35\%} & \textbf{54.2$\pm$2.36\%} & \textbf{49.6$\pm$4.21\%} & \textbf{36.4$\pm$2.60\%} & \textbf{100.0$\pm$0.00\%} & \textbf{71.5$\pm$2.57\%} & \textbf{57.2$\pm$3.79\%} & \textbf{45.6$\pm$1.93\%} \\
\bottomrule
	\end{tabular}}
	\label{aoaresult}
\end{table*}

\begin{table*}[htbp]
	\caption{\changedtext{Error Rate (Top-1) of Transfer Attacks on ResNet50}}
	\centering
	\begin{tabular}{r|cccccccc}
\toprule
Method & DN121 \cite{huang2017densely} & IncRNV2 \cite{szegedy2017inception} & IncV3 \cite{szegedy2016rethinking}  & NASNetL \cite{zoph2018learning} & RN152 \cite{he2016deep} & RN50 \cite{he2016deep} & VGG19 \cite{simonyan2014very} & Xception \cite{chollet2017xception}\\ \midrule
CW & 66.6$\pm$1.24\% & 37.9$\pm$4.77\% & 39.6$\pm$2.92\% & 28.8$\pm$2.58\% & 47.3$\pm$4.69\% & 100.0$\pm$0.00\% & 54.2$\pm$4.27\% & 37.4$\pm$2.67\% \\
MI-DI-CW & 66.9$\pm$1.91\% & 39.4$\pm$4.03\% & 42.9$\pm$1.59\% & 32.3$\pm$3.83\% & 50.2$\pm$4.74\% & 99.8$\pm$0.24\% & 57.9$\pm$3.40\% & 39.9$\pm$2.92\% \\
MI-TI-CW & 63.4$\pm$3.35\% & 42.0$\pm$3.33\% & 44.6$\pm$1.02\% & 33.7$\pm$1.96\% & 51.6$\pm$3.77\% & 99.7$\pm$0.24\% & 60.2$\pm$2.80\% & 40.6$\pm$2.40\% \\

SI-CW & 80.3$\pm$1.86\% & 46.4$\pm$2.22\% & 51.6$\pm$2.60\% & 38.3$\pm$3.53\% & 63.9$\pm$1.50\% & 99.9$\pm$0.20\% & 66.5$\pm$1.67\% & 48.8$\pm$3.70\% \\ \midrule
PGD & 67.8$\pm$1.83\% & 35.6$\pm$4.21\% & 38.7$\pm$2.25\% & 28.4$\pm$3.17\% & 46.8$\pm$3.71\% & 100.0$\pm$0.00\% & 54.2$\pm$2.56\% & 37.4$\pm$4.08\% \\
MI-DI-PGD & 70.5$\pm$1.30\% & 43.3$\pm$3.33\% & 45.8$\pm$2.58\% & 35.7$\pm$3.53\% & 55.9$\pm$3.68\% & 99.5$\pm$0.00\% & 62.1$\pm$1.93\% & 43.3$\pm$2.42\% \\
MI-TI-PGD & 68.6$\pm$0.97\% & 44.6$\pm$2.18\% & 49.5$\pm$1.30\% & 38.0$\pm$1.00\% & 54.2$\pm$1.99\% & 99.3$\pm$0.51\% & 64.2$\pm$2.29\% & 45.3$\pm$1.72\% \\

SI-PGD & 81.2$\pm$1.63\% & 48.7$\pm$1.91\% & 53.0$\pm$0.95\% & 38.6$\pm$2.06\% & 66.1$\pm$2.46\% & 100.0$\pm$0.00\% & 69.5$\pm$2.10\% & 49.1$\pm$1.59\% \\ \midrule
AoA & 78.4$\pm$2.44\% & 49.0$\pm$1.87\% & 52.2$\pm$2.66\% & 39.6$\pm$3.61\% & 63.4$\pm$2.63\% & 99.9$\pm$0.20\% & 65.6$\pm$2.82\% & 51.1$\pm$2.18\% \\
MI-DI-AoA & 74.1$\pm$1.02\% & 50.4$\pm$2.92\% & 52.0$\pm$3.32\% & 44.2$\pm$3.39\% & 58.7$\pm$3.59\% & 99.8$\pm$0.24\% & 66.4$\pm$4.20\% & 50.6$\pm$3.01\% \\
MI-TI-AoA & 79.2$\pm$1.21\% & 58.7$\pm$4.27\% & 62.5$\pm$3.52\% & 52.2$\pm$3.23\% & 67.5$\pm$2.76\% & 99.8$\pm$0.40\% & 75.3$\pm$2.89\% & 58.9$\pm$1.56\% \\

SI-AoA & \textbf{90.5$\pm$0.89\%} & \textbf{64.6$\pm$2.71\%} & \textbf{66.1$\pm$3.89\%} & \textbf{57.9$\pm$2.20\%} & \textbf{78.8$\pm$1.75\%} & \textbf{100.0$\pm$0.00\%} & \textbf{80.4$\pm$2.73\%} & \textbf{64.6$\pm$3.07\%} \\
\bottomrule
	\end{tabular}
	\label{aoaresult2}
\end{table*}

\subsection{AoA under Defenses}
Our main contribution in this paper is for black-box attack by increasing the transferability. It is not necessary that AoA can break defenses, but indeed, it is interesting to evaluate the attack performance under several defenses. In this experiment, we apply PGD, CW, and AoA, all enhanced by SI 
to attack ResNet50.
We consider defenses that have been verified effective on ImageNet \cite{carlini2017adversarial}. Those defense methods can be roughly categorized as preprocessing-based and adversarial-training-based, which could be used together.

Preprocessing-based defenses are to eliminate the adversarial perturbation. We use JPEG Compression \cite{liu2018feature}, Pixel Deflection \cite{prakash2018deflecting}, Total Variance Minimization (TVM) \cite{guo2017countering} with provided parameters. Another idea is to add the randomness to observe the variance of the outputs. For example, Random Smoothing \cite{cohen2019certified} makes prediction by $m$ intermediate images, which is crafted by Gaussian noise from the input image. We choose $m=100$ and the Gaussian noise scale $\sigma=0.25*255$ here.

Adversarial training is to re-train the neural networks by adversarial samples. In \cite{kurakin2016adversarial}, InceptionV3adv and InceptionResNetV2adv are designed and \cite{xie2019feature} proposes ResNetXt101denoise with denoising blocks in architectures to secure the model.

\lchanged{M05}{\link{R0.0} \link{R3.1}}{Table \ref{defenses}} gives the comprehensive black-box attack performance under defenses. Generally speaking, the preprocessing-based defenses decrease the error rate for about 5\% to 10\% and SI-AoA maintains the highest transfer rate. Adversarial-trained models (indicated by underlines in tables) exhibit a strong robustness to attacks, including SI-AoA (but still, it is better than SI-PGD, SI-CW). That means although samples generated by SI-AoA are different to others, the distribution can still be captured by adversarial training. Developing adversarial attacks that can defeat adversarial training is interesting but out of our scope. Random smoothing generally has a low error rate but its inference time is much longer than other methods, generally $m$ times and hence it is not a fair comparison. In our experiment, random smoothing seems not to work well on adversarial-trained models, sometimes even oppositely, which is also interesting but in the field of defenses.

\begin{table*}[htbp]
	\caption{\changedtext{Error Rate (Top-1) under Defenses (ResNet50 as the surrogate model)}}
	\centering
	\begin{tabular}{r|l|cccccc}
    \toprule
Victim & Method & None & JPEG \cite{liu2018feature} & Pixel \cite{prakash2018deflecting} & Random \cite{xie2017mitigating} & TVM \cite{guo2017countering} & Smooth \cite{cohen2019certified} \\ \midrule
 & SI-CW & 80.3$\pm$1.86\% & 64.9$\pm$2.40\% & 67.2$\pm$2.20\% & 64.5$\pm$3.99\% & 70.2$\pm$1.63\% & 60.0$\pm$2.26\% \\
DN121 \cite{huang2017densely} & SI-PGD & 81.2$\pm$1.63\% & 65.1$\pm$1.24\% & 66.4$\pm$0.58\% & 64.0$\pm$3.44\% & 69.7$\pm$1.29\% & 60.0$\pm$2.26\% \\
 & SI-AoA & \textbf{90.5$\pm$0.89\%} & \textbf{81.0$\pm$3.32\%} & \textbf{82.1$\pm$2.85\%} & \textbf{78.0$\pm$3.70\%} & \textbf{83.7$\pm$3.14\%} & \textbf{63.4$\pm$2.35\%} \\ \midrule

 & SI-CW & 46.4$\pm$2.22\% & 38.0$\pm$2.17\% & 38.3$\pm$0.93\% & 40.3$\pm$3.04\% & 41.0$\pm$1.64\% & 31.7$\pm$2.19\% \\
IncRNV2 \cite{szegedy2017inception} & SI-PGD & 48.7$\pm$1.91\% & 39.8$\pm$0.93\% & 39.3$\pm$0.75\% & 40.0$\pm$3.11\% & 42.1$\pm$0.86\% & 31.8$\pm$1.70\% \\
 & SI-AoA & \textbf{64.6$\pm$2.71\%} & \textbf{56.7$\pm$1.72\%} & \textbf{58.2$\pm$3.91\%} & \textbf{57.8$\pm$4.37\%} & \textbf{59.5$\pm$2.63\%} & \textbf{34.6$\pm$3.24\%} \\ \midrule

 & SI-CW & 51.6$\pm$2.60\% & 43.2$\pm$3.39\% & 42.7$\pm$2.98\% & 46.2$\pm$2.34\% & 46.1$\pm$3.47\% & 33.5$\pm$4.73\% \\
IncV3 \cite{szegedy2016rethinking} & SI-PGD & 53.0$\pm$0.95\% & 44.8$\pm$3.33\% & 45.0$\pm$2.98\% & 47.9$\pm$3.09\% & 48.3$\pm$3.23\% & 32.6$\pm$5.66\% \\
 & SI-AoA & \textbf{66.1$\pm$3.89\%} & \textbf{62.3$\pm$3.87\%} & \textbf{62.4$\pm$4.12\%} & \textbf{62.9$\pm$2.67\%} & \textbf{64.1$\pm$3.79\%} & \textbf{37.5$\pm$6.18\%} \\ \midrule

 & SI-CW & 38.3$\pm$3.53\% & 31.3$\pm$3.09\% & 32.4$\pm$4.12\% & 35.2$\pm$2.93\% & 34.0$\pm$4.57\% & 23.7$\pm$3.68\% \\
 NASNetL \cite{zoph2018learning} & SI-PGD & 38.6$\pm$2.06\% & 30.8$\pm$3.59\% & 31.5$\pm$2.92\% & 34.3$\pm$4.07\% & 34.6$\pm$2.96\% & 23.5$\pm$3.35\% \\
 & SI-AoA & \textbf{57.9$\pm$2.20\%} & \textbf{49.2$\pm$3.71\%} & \textbf{53.0$\pm$4.01\%} & \textbf{52.7$\pm$3.93\%} & \textbf{53.0$\pm$3.32\%} & \textbf{29.3$\pm$2.80\%} \\ \midrule

 & SI-CW & 63.9$\pm$1.50\% & 51.4$\pm$1.91\% & 51.6$\pm$1.85\% & 48.9$\pm$3.85\% & 56.6$\pm$1.56\% & 41.2$\pm$5.28\% \\
RN152 \cite{he2016deep}  & SI-PGD & 66.1$\pm$2.46\% & 52.8$\pm$2.56\% & 54.1$\pm$1.53\% & 51.5$\pm$3.39\% & 58.4$\pm$1.83\% & 40.2$\pm$4.81\% \\
 & SI-AoA & \textbf{78.8$\pm$1.75\%} & \textbf{70.3$\pm$3.56\%} & \textbf{72.8$\pm$4.49\%} & \textbf{67.1$\pm$2.82\%} & \textbf{75.6$\pm$3.93\%} & \textbf{44.2$\pm$5.07\%} \\ \midrule

& SI-CW & 99.9$\pm$0.20\% & 98.5$\pm$0.84\% & 98.7$\pm$0.81\% & 89.5$\pm$2.59\% & 99.6$\pm$0.49\% & 93.4$\pm$0.94\% \\
RN50 \cite{he2016deep} & SI-PGD & 100.0$\pm$0.00\% & 99.1$\pm$0.49\% & 99.4$\pm$0.58\% & 90.8$\pm$1.33\% & 99.6$\pm$0.37\% & 92.4$\pm$1.71\% \\
& SI-AoA & \textbf{100.0$\pm$0.00\%} & \textbf{99.9$\pm$0.20\%} & \textbf{99.8$\pm$0.40\%} & \textbf{95.6$\pm$2.13\%} & \textbf{99.9$\pm$0.20\%} & \textbf{94.1$\pm$1.20\%} \\ \midrule

 & SI-CW & 66.5$\pm$1.67\% & 60.7$\pm$4.27\% & 60.6$\pm$3.20\% & 62.9$\pm$4.07\% & 63.3$\pm$5.09\% & 89.8$\pm$1.89\% \\
VGG19 \cite{simonyan2014very}  & SI-PGD & 69.5$\pm$2.10\% & 62.8$\pm$3.54\% & 61.4$\pm$4.92\% & 65.7$\pm$3.80\% & 65.2$\pm$4.25\% & 89.6$\pm$1.73\% \\
 & SI-AoA & \textbf{80.4$\pm$2.73\%} & \textbf{77.7$\pm$4.43\%} & \textbf{78.5$\pm$3.77\%} & \textbf{77.1$\pm$4.52\%} & \textbf{79.8$\pm$4.04\%} & \textbf{89.9$\pm$2.18\%} \\ \midrule

 & SI-CW & 48.8$\pm$3.70\% & 40.6$\pm$3.81\% & 40.9$\pm$3.71\% & 44.0$\pm$2.92\% & 44.7$\pm$3.23\% & 36.5$\pm$4.38\% \\
Xception \cite{chollet2017xception} & SI-PGD & 49.1$\pm$1.59\% & 40.8$\pm$3.59\% & 43.0$\pm$4.02\% & 43.5$\pm$3.89\% & 44.7$\pm$3.37\% & 37.1$\pm$3.35\% \\
 & SI-AoA & \textbf{64.6$\pm$3.07\%} & \textbf{57.6$\pm$3.26\%} & \textbf{58.4$\pm$1.80\%} & \textbf{61.1$\pm$3.89\%} & \textbf{59.0$\pm$2.65\%} & \textbf{40.9$\pm$4.52\%} \\ \midrule

 & SI-CW & 31.2$\pm$1.29\% & 33.8$\pm$2.50\% & 35.0$\pm$3.35\% & 38.1$\pm$3.73\% & 37.0$\pm$4.27\% & 96.5$\pm$1.44\% \\
\underline{IncV3adv} \cite{kurakin2016adversarial}  & SI-PGD & 31.5$\pm$3.08\% & 34.3$\pm$3.44\% & 35.8$\pm$2.99\% & 39.2$\pm$3.14\% & 38.4$\pm$2.85\% & 96.2$\pm$1.13\% \\
 & SI-AoA & \textbf{53.7$\pm$2.25\%} & \textbf{52.7$\pm$2.20\%} & \textbf{54.9$\pm$3.15\%} & \textbf{55.1$\pm$2.78\%} & \textbf{56.2$\pm$2.71\%} & \textbf{96.2$\pm$1.16\%} \\ \midrule

 & SI-CW & 26.4$\pm$1.59\% & 27.4$\pm$2.03\% & 27.6$\pm$2.63\% & 30.1$\pm$4.78\% & 28.2$\pm$3.66\% & 81.7$\pm$3.74\% \\
\underline{IncRNV2adv} \cite{kurakin2016adversarial}  & SI-PGD & 26.1$\pm$1.98\% & 27.9$\pm$0.86\% & 28.5$\pm$2.51\% & 29.7$\pm$3.64\% & 29.8$\pm$0.93\% & 81.5$\pm$3.47\% \\
 & SI-AoA & \textbf{44.0$\pm$1.52\%} & \textbf{44.2$\pm$3.23\%} & \textbf{46.2$\pm$3.71\%} & \textbf{48.0$\pm$4.55\%} & \textbf{47.0$\pm$2.30\%} & \textbf{82.3$\pm$3.16\%} \\ \midrule

 & SI-CW & 18.0$\pm$3.13\% & 18.2$\pm$3.11\% & 18.2$\pm$3.33\% & 44.4$\pm$3.69\% & 18.1$\pm$3.22\% & 70.4$\pm$2.26\% \\
\underline{RNXt101den} \cite{xie2019feature}  & SI-PGD & 18.2$\pm$2.87\% & 18.5$\pm$2.88\% & 18.9$\pm$3.17\% & 44.6$\pm$3.46\% & 18.4$\pm$3.31\% & 70.5$\pm$2.09\% \\
 & SI-AoA & \textbf{18.7$\pm$3.01\%} & \textbf{19.2$\pm$2.71\%} & \textbf{19.1$\pm$2.97\%} & \textbf{44.6$\pm$3.48\%} & \textbf{19.0$\pm$2.88\%} & \textbf{70.5$\pm$2.26\%} \\
\bottomrule
	\end{tabular}
	\label{defenses}
\end{table*}

\subsection{DAmageNet}
The above experiments verify that AoA has a promising transferability, which then makes it possible to generate adversarial samples that are able to beat many well-trained DNNs. An adversarial dataset will be very useful for evaluating robustness and defense methods. To establish an adversarial dataset, we use SI-AoA to attack VGG19 to generate samples from all 50000 samples from ImageNet validation set. Since the original images come from ImageNet training set and the adversarial samples are going to cheat neural networks, we hence name this dataset as DAmageNet.

\begin{figure*}[htbp]
	\centering
	\includegraphics[width=\hsize]{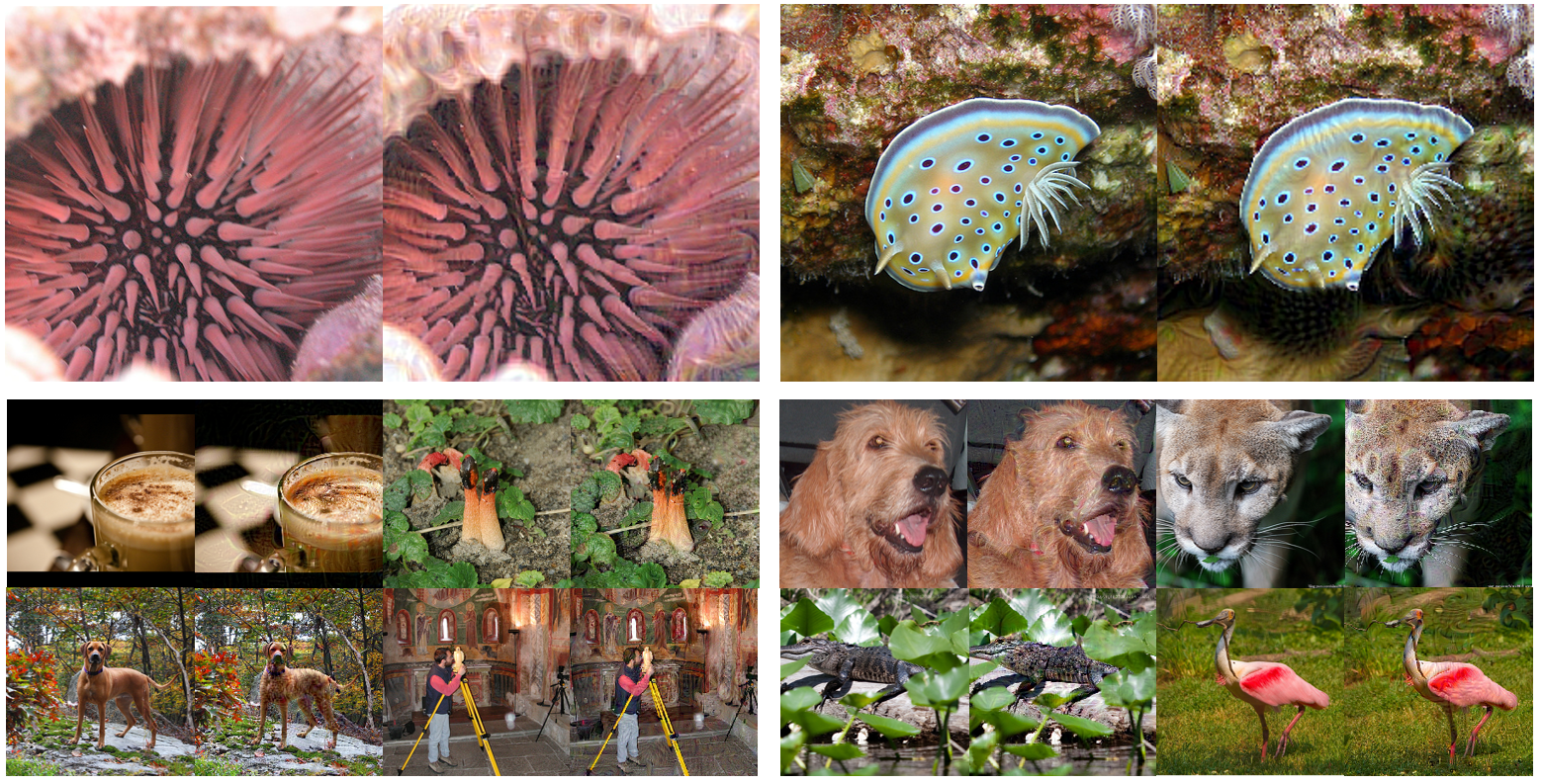}
	\caption{Samples in ImageNet and DAmageNet. The images on the left are original samples from ImageNet. The images on the right are adversarial samples from DAmageNet. One could observe that these images look similar and human beings have no problem to recognize them as the same class.}
	\label{samples}
\end{figure*}

DAmageNet contains 50000 adversarial samples and could be downloaded from \url{http://www.pami.sjtu.edu.cn/Show/56/122}. The samples are named the same as the original ones in ImageNet validation set. Accordingly, users could easily find the corresponding samples as well as their labels. The average RMSE between samples in DAmageNet and those in ImageNet is 7.23. In Fig. \ref{samples}, we show several image pairs in ImageNet and DAmageNet.

\begin{table*}[htpb]
	\caption{Error Rate (Top-1) on ImageNet and DAmageNet}
	\centering
	\begin{tabular}{r|cc|cccc}
    \toprule
 & \multicolumn{2}{c}{No defense} & \multicolumn{4}{c}{Defenses on DAmageNet} \\
Victim & ImageNet \cite{deng2009imagenet}& DAmageNet & JPEG \cite{liu2018feature}& Pixel \cite{prakash2018deflecting}& Random \cite{xie2017mitigating}& TVM \cite{guo2017countering}\\ \midrule
VGG16 \cite{simonyan2014very} & 38.51  & 99.85  & 99.67  & 99.70  & 99.19  & 99.76 \\
VGG19 \cite{simonyan2014very} & 38.60  & 99.99  & 99.99  & 99.99  & 99.96  & 99.99\\
RN50 \cite{he2016deep}& 36.65  & 93.94  & 91.88  & 92.48  & 92.52  & 93.08\\
RN101 \cite{he2016deep}& 29.38  & 88.13  & 85.44  & 86.23  & 86.12  & 87.06 \\
RN152 \cite{he2016deep}& 28.65  & 86.78  & 83.93  & 84.83  & 84.71  & 85.68\\
NASNetM \cite{zoph2018learning}& 27.03  & 92.81  & 90.42  & 91.43  & 90.31  & 91.86 \\
NASNetL \cite{zoph2018learning}& 17.77  & 86.32  & 83.31  & 84.87  & 84.91 & 85.53 \\
IncV3 \cite{szegedy2016rethinking} & 22.52  & 89.84  & 87.82  & 89.01  & 88.49 & 89.59  \\
IncRNV2 \cite{szegedy2017inception}& 24.60  & 88.09  & 85.01  & 85.95  & 89.04 & 86.79 \\
Xception \cite{chollet2017xception} & 21.38  & 90.57  & 88.53  & 89.77  & 86.03 & 90.32 \\
DN121 \cite{huang2017densely}& 26.85  & 96.14  & 93.96  & 94.85  & 93.82  & 95.30\\
DN169 \cite{huang2017densely}& 25.16  & 94.09  & 91.72  & 92.78  & 91.78  & 93.36 \\
DN201 \cite{huang2017densely}& 24.36  & 93.44  & 90.52  & 91.71  & 90.86  & 92.45 \\
\underline{IncV3adv} \cite{kurakin2016adversarial}& 22.86  & 82.23  & 82.03  & 83.35  & 82.88 & 83.95 \\
\underline{IncV3advens3} \cite{tramer2017ensemble}& 24.12  & 80.72  & 80.35  & 81.68  & 81.57  & 82.36 \\
\underline{IncV3advens4}  \cite{tramer2017ensemble}& 24.45  & 79.26  & 78.86  & 79.96  & 79.76  & 80.8 \\
\underline{IncRNV2adv} \cite{kurakin2016adversarial}& 20.03  & 76.42  & 75.71  & 76.85  & 76.86 & 77.73 \\
\underline{IncRNV2advens} \cite{tramer2017ensemble}& 20.35  & 70.70  & 71.09  & 72.32  & 73.32  & 73.04 \\
\underline{RNXt101den} \cite{xie2019feature}& 32.20  & 35.40  & 36.27  & 36.65  & 55.53  & 36.21\\
\bottomrule
	\end{tabular}
	\label{rate}
\end{table*}

To the best of our knowledge, DAmageNet is the first adversarial dataset, which can be used to evaluate model robustness and defenses. As an example, we use several well-trained models to recognize the images in DAmageNet. Several neural networks strengthened by adversarial training are considered as well. The error rate (top-1) is reported in Table \ref{rate}. The models are from Keras Application and the test error may differ from original references. One could observe that i) all the listed 13 undefended models are not robust: DAmageNet increases the error rate of all 13 undefended models to over 85\%; ii) the 5 listed adversarial-trained models have a slightly better performance and the error rate is over $70\%$; iii) DAmageNet resists 4 tested defenses with almost no drop on the error rate compared to other methods; iv) feature denoising model shows promising robustness but simply combining it with preprocessing-based defence does not work well.

\section{Conclusion}\label{conclusion}
To improve the transferability of adversarial attack, we are the first to attack on attention and achieve a great performance on the black-box attack. The high transferability of AoA relies on the semantic features shared by different DNNs. AoA enjoys a significant increase in transferability when the traditional cross entropy loss is replaced with the attention loss. Since AoA alters the loss only, it could be easily combined with other transferability-enhancement methods, e.g., SI \cite{lin2019nesterov}, and achieve a state-of-the-art performance.

By SI-AoA, we generate DAmageNet, the first dataset containing samples with a small perturbation and a high transfer rate (an error rate over 85\% for undefended models and over 70\% for adversarial-trained models). DAmageNet provides a benchmark to evaluate the robustness of DNNs by elaborately-crafted adversarial samples.

AoA has found the common vulnerability of DNNs in attention. Also, attention is just one semantic feature and attacking on other semantic features shared by DNNs is also promising to have good transferability.
\ifCLASSOPTIONcompsoc
  \section*{Acknowledgments}
\else
  \section*{Acknowledgment}
\fi
This work was partially supported by National Key Research Development Project (No. 2018AAA0100702, 2019YFB1311503) and National Natural Science Foundation of China (No. 61977046, 61876107, U1803261).

The authors are grateful to the anonymous reviewers for their insightful comments.
\ifCLASSOPTIONcaptionsoff
  \newpage
\fi

\bibliographystyle{IEEEtran}
\bibliography{Reference}

\vspace{-1cm}
\begin{IEEEbiography}[{\includegraphics[width=1in,height=1.25in,clip,keepaspectratio]{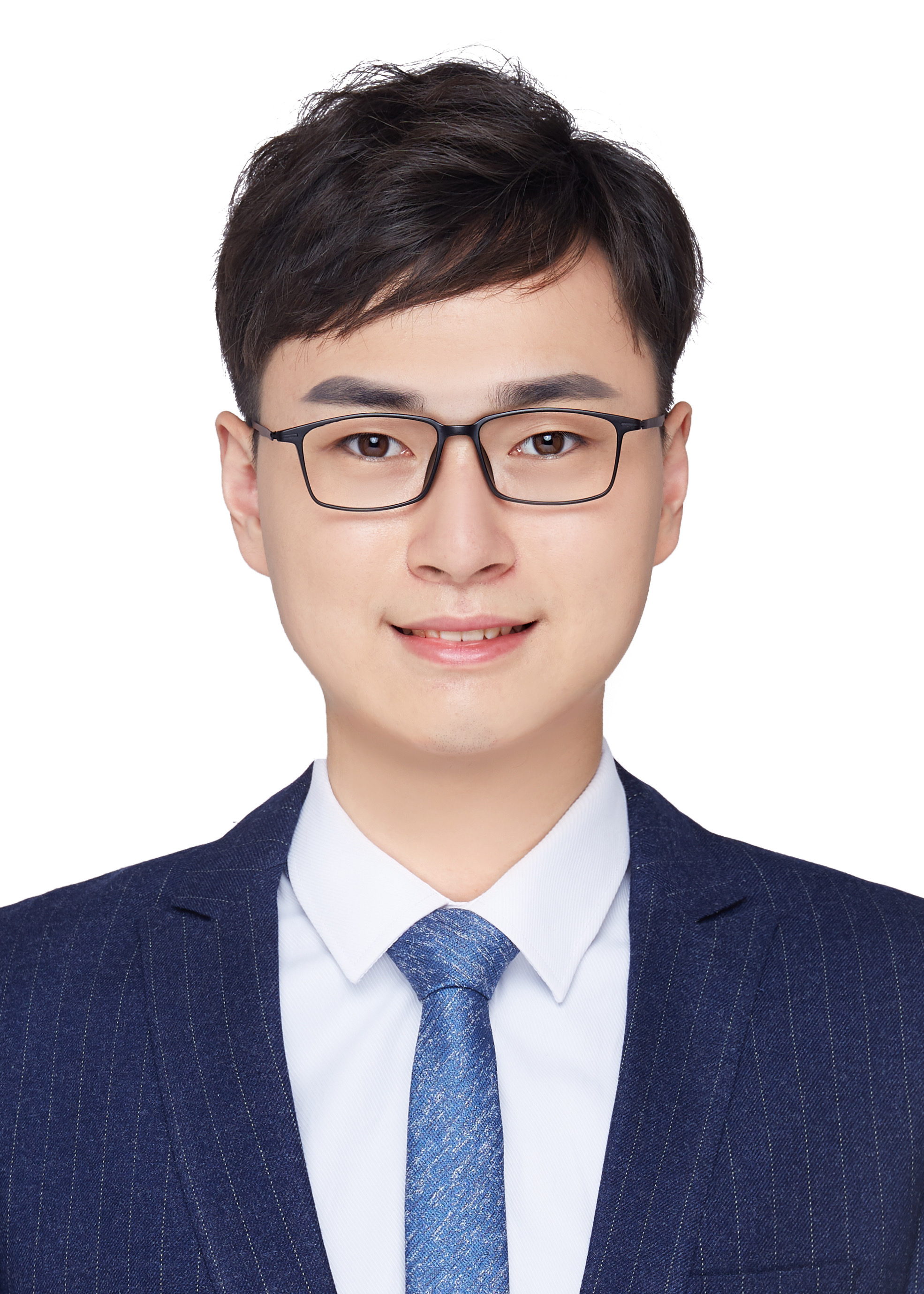}}]{Sizhe Chen}
received his B.S. degree in Shanghai Jiao Tong University, Shanghai, China, in 2020. He is now a master student at the Institute of Image Processing and Pattern Recognition, Shanghai Jiao Tong University, Shanghai, China. His research interests are model security, robust learning, and interpretability of DNN.
\end{IEEEbiography}

\begin{IEEEbiography}[{\includegraphics[width=1in,height=1.25in,clip,keepaspectratio]{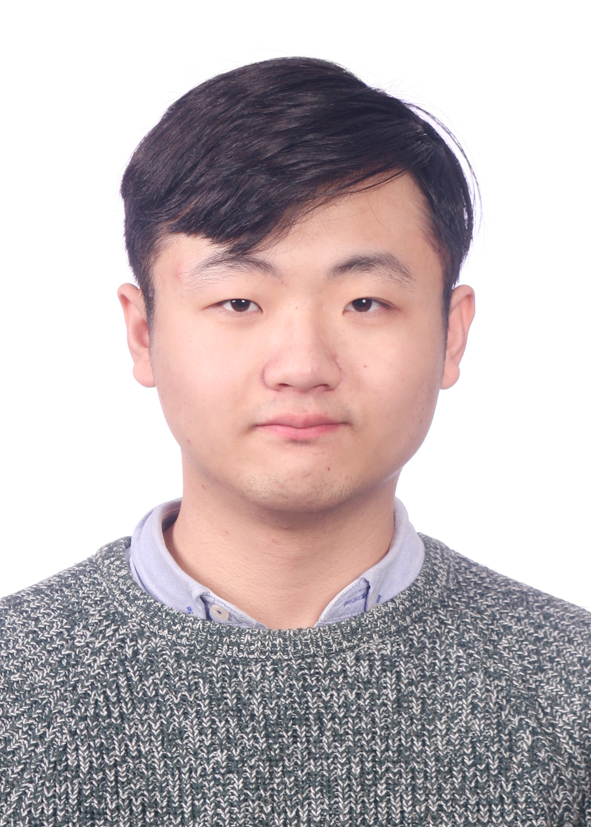}}]{Zhengbao He}
	is a senior student in Department of Automation, Shanghai Jiao Tong University, Shanghai, China. He is now doing research at the Institute of Image Processing and Pattern Recognition, Shanghai Jiao Tong University. His research interests are adversarial attack and deep learning.
\end{IEEEbiography}

\begin{IEEEbiography}[{\includegraphics[width=1in,height=1.25in,clip,keepaspectratio]{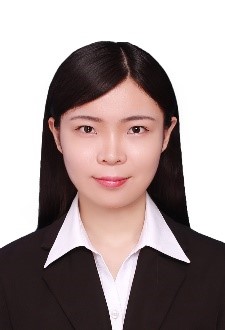}}]{Chengjin Sun}
	received her B.S. degree in Nanjing University, Nanjing, China, in 2018. She is now a master student at the Institute of Image Processing and Pattern Recognition, Shanghai Jiao Tong University, Shanghai, China. Her research interests are adversarial robustness for deep learning.
\end{IEEEbiography}

\begin{IEEEbiography}[{\includegraphics[width=1in,height=1.25in,clip,keepaspectratio]{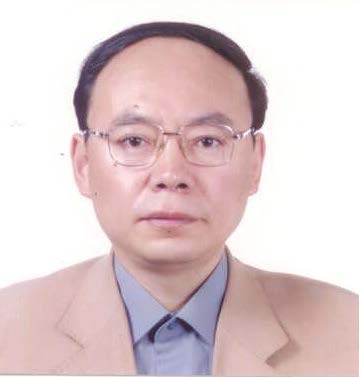}}]{Jie Yang}
received his Ph.D. from the Department of Computer Science, Hamburg University, Hamburg, Germany,
in 1994. Currently, he is a professor at the Institute of Image Processing and Pattern
recognition, Shanghai Jiao Tong University, Shanghai, China. He has led many research projects (e.g.,
National Science Foundation, 863 National High Technique Plan), had one book published in Germany,
and authored more than 300 journal papers. His major research interests are object detection and
recognition, data fusion and data mining, and medical image processing.
\end{IEEEbiography}

\begin{IEEEbiography}[{\includegraphics[width=1in,height=1.25in,clip,keepaspectratio]{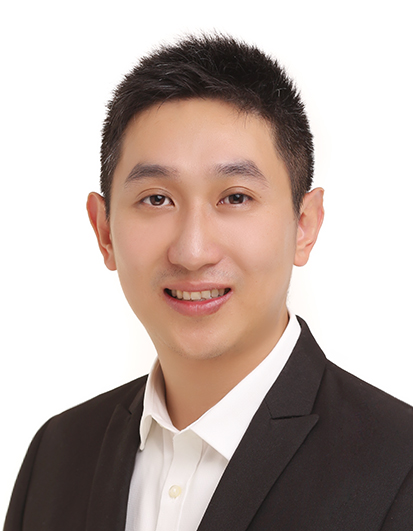}}]{Xiaolin Huang}
(S'10-M'12-SM'18) received the B.S. degree in control science and engineering, and the B.S. degree in applied mathematics from Xi'an Jiaotong University, Xi'an, China in 2006. In 2012, he received the Ph.D. degree in control science and engineering from Tsinghua University, Beijing, China. From 2012 to 2015, he worked as a postdoctoral researcher in ESAT-STADIUS, KU Leuven, Leuven, Belgium. After that he was selected as an Alexander von Humboldt Fellow and working in Pattern Recognition Lab, the Friedrich-Alexander-Universit\"{a}t Erlangen-N\"{u}rnberg, Erlangen, Germany. From 2016, he has been an Associate Professor at Institute of Image Processing and Pattern Recognition, Shanghai Jiao Tong University, Shanghai, China. In 2017, he was awarded by "1000-Talent Plan" (Young Program).

His current research areas include machine learning and optimization, especially for robustness and sparsity of both kernel learning and deep neural networks.
\end{IEEEbiography}

\end{document}